\documentclass{article}
\usepackage{spconf,amsmath,graphicx}

\usepackage{microtype}
\usepackage{booktabs}
\usepackage{amsfonts,amssymb}
\usepackage{multirow}
\usepackage{stmaryrd}
\usepackage{stfloats}
\usepackage{xspace}
\usepackage{color}
\usepackage{enumitem}

\newcommand{\ie}{\emph{i.e.,}\xspace}
\newcommand{\eg}{\emph{e.g.,}\xspace}

\title{An Embarrassingly Simple Model for Dialogue Relation Extraction}

\name{Fuzhao Xue$^{1}$, Aixin Sun$^{1}$, Hao Zhang$^{1,2}$, Jinjie Ni$^{1}$, Eng-Siong Chng$^{1}$
\thanks{This research is supported by the Agency for Science, Technology and Research (A*STAR) under its AME Programmatic Funding Scheme (Project \#A19E2b0098) and the National Research Foundation Singapore under its AI Singapore Programme (Award Number: AISG-100E-2018-006).}
}
\address{$^{1}$School of Computer Science and Engineering, Nanyang Technological University, Singapore \\
$^{2}$Centre for Frontier AI Research, Agency for Science, Technology and Research, Singapore
}

\begin{document}

\maketitle

\begin{abstract}
Dialogue relation extraction (RE) is to predict the relation type of two entities mentioned in a dialogue. In this paper, we propose a simple yet effective model named SimpleRE for the RE task. SimpleRE captures the interrelations among multiple relations in a dialogue through a novel input format named BERT Relation Token Sequence (BRS). In BRS, multiple [CLS] tokens are used to capture possible relations between different pairs of entities mentioned in the dialogue. A Relation Refinement Gate (RRG) is then designed to extract relation-specific semantic representation in an adaptive manner. Experiments on the DialogRE dataset show that SimpleRE achieves the best performance, with much shorter training time. Further, SimpleRE outperforms all direct baselines on sentence-level RE without using external resources. 
\end{abstract}


%===========================
\section{Introduction}\label{sec:intro}
%===========================
Relation extraction (RE) is to identify the semantic relation type between two entities mentioned in a piece of text, \eg a sentence or a dialogue. Table~\ref{tbl-dialog-example} shows an example dialogue. The RE task is to predict the relation type of a pair of entities like ``Monica'' and ``S2'' (\ie an argument pair)  mentioned in the dialogue, from a set of predefined relations. Researchers have tried to improve Dialogue RE by considering speaker information~\cite{yu-etal-2020-dialogue} or trigger tokens~\cite{xue2020gdpnet}. There are also solutions based on graph attention network, where a graph models speaker, entity, entity-type, and utterance nodes~\cite{chen2020dialogue}. However, Transformer-based models remain strong competitors~\cite{yu-etal-2020-dialogue,xue2020gdpnet}.

A dialogue may mention multiple pairs of entities, reflected by  annotations in the DialogRE dataset~\cite{yu-etal-2020-dialogue} (see Table~\ref{tbl-dialog-example}). Among  multiple pairs of entities, the relations mentioned in the same dialog often interrelate with each other to some extent. An example is shown in Table~\ref{tbl-dialog-example}, ``Richard'' and ``Monica'' in the first few utterances show two possible relations, \ie ``positive\_impression'' or ``girl/boyfriend''. The last utterance indicates that ``Monica'' is girlfriend of ``S2''; hence ``Richard'' and ``Monica'' can only be related by ``positive\_impression''. We argue that such interrelationships could be helpful for relation extraction. 

\begin{table}[t]
\centering
\caption{An example from DialogRE dataset~\cite{yu-etal-2020-dialogue}. Relations of two pairs of entities are annotated. }
\begin{tabular}{@{}lp{2.57in}}
\toprule
\textbf{S1}: & Where the hell have you been?! \\
\textbf{S2}: & I was making a coconut phone with the professor.\\
\textbf{S1}: & Richard told Monica he wants to marry her!    \\
\textbf{S2}: & What?!     \\
\textbf{S1}: & Yeah! Yeah, I’ve been trying to find ya to tell to stop messing with her and maybe I would have if these damn boat shoes wouldn’t keep flying off!    \\
\textbf{S2}: & My—Oh my God! \\
\textbf{S1}: & I know! They suck!!      \\
\textbf{S2}: & He’s not supposed to ask my girlfriend to marry him! I’m supposed to do that!     \\
\bottomrule
\end{tabular}
\begin{tabular}{@{}llp{1.5in}}
 & \textbf{Argument pair}      & \textbf{Relation type}     \\
\textbf{R1}  & (Monica, S2)       & girl/boyfriend             \\
\textbf{R2}  & (Richard, Monica)  & positive\_impression       \\
\bottomrule
\end{tabular}

\label{tbl-dialog-example}
\end{table}

\begin{figure*}
\centering
\caption{The architecture of SimpleRE. An entity may contain one or more tokens as illustrated.}
\includegraphics[width=0.9\textwidth]{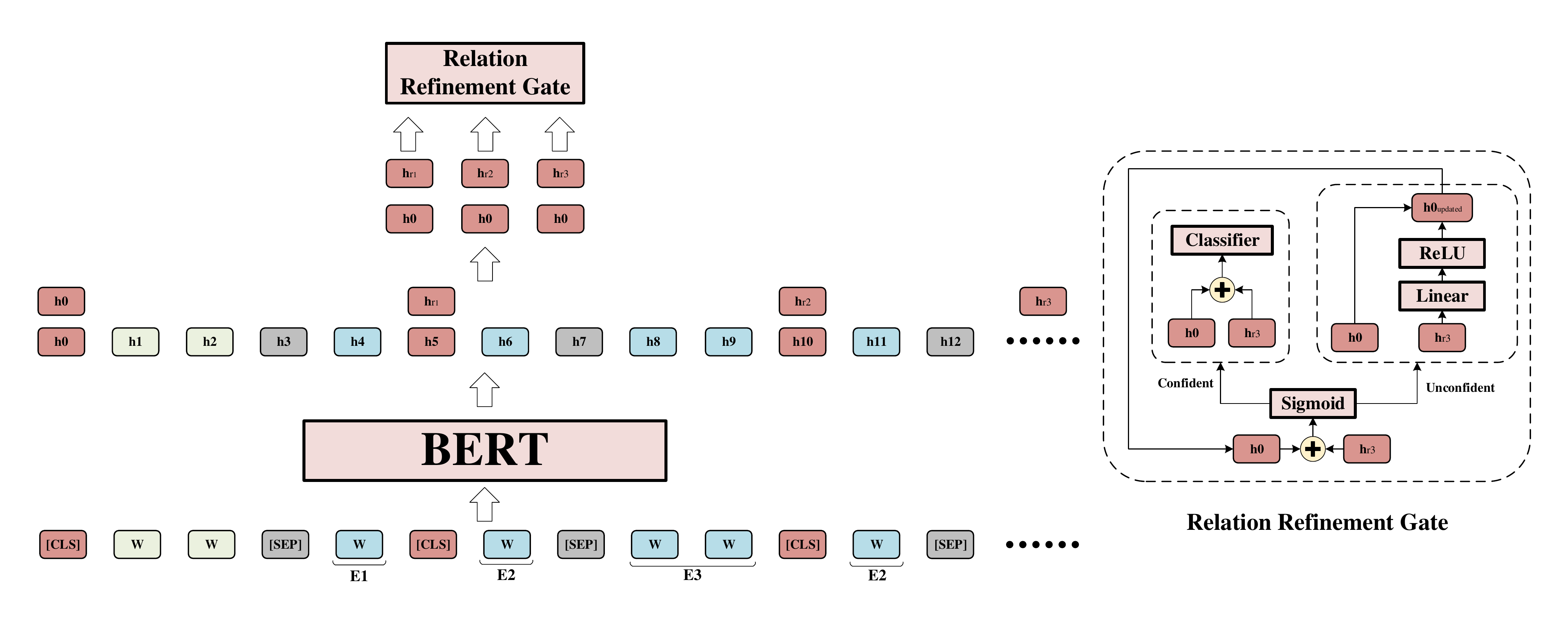}

\label{fig:overview}
\vspace{-2.0em}
\end{figure*}

In this paper, we propose SimpleRE, an extremely simple model, to reason and learn interrelations among tokens and relations. SimpleRE is built on top of BERT. Due to its strong modeling capability, BERT is the natural choice to model such interrelationships. We first design a BERT Relation Token Sequence (BRS). BRS contains multiple ``[CLS]'' tokens in input sequence, with the aim to capture relations between multiple pairs of entities. We then propose a Relation Refinement Gate (RRG) to refine the semantic representation of each relation for target relation prediction in an adaptive manner. 

On the DialogRE dataset, SimpleRE achieves  best $F1$ over two BERT-based methods, BERTs~\cite{yu-etal-2020-dialogue} and GDPNet~\cite{xue2020gdpnet}, by a large margin. As a simple model, the training of SimpleRE is at least 5 times faster than these two models. We also show that BRS is effective on sentence-level RE, and the adapted SimpleRE beats all direct baselines on the TACRED dataset. 

%===========================
\section{S\lowercase{imple}RE}\label{sec:method}
%===========================
The  architecture of SimpleRE is shown in Figure~\ref{fig:overview}. Its novelty are two-fold: (i) BERT Relation Token Sequence (BRS), \ie the input format to BERT,  and (ii) Relation Refinement Gate (RRG), \ie the way to utilize BERT encoding. 
 
%===========================
\subsection{Problem formulation}
%===========================
Let $\mathcal{R}$ be a set of predefined relation types. Let $X=\{x_1,x_2,\dots,x_T\}$ be a text sequence with $T$ tokens, where $x_t$ is the token at $t$-th position. $X$ denotes an entire dialogue for Dialogue RE, or a single sentence for sentence-level RE. Between $n$ pairs of entities mentioned in $X$, there could be multiple relations $R=\{r_1,r_2,\dots,r_n\}$. The $i$-th relation $r_i\in \mathcal{R}$ is predicted for an argument pair: \textit{subject entity} $E_s^i$ and \textit{object entity} $E_o^i$. Note that, an entity may contain one or more tokens. In this problem setting, the pairs of entities whose relations are to be predicted are known.  

%===========================
\subsection{BERT Relation Token Sequence}
%===========================
BERT~\cite{devlin-etal-2019-bert} based models are powerful in modeling semantics in text sequences~\cite{10.1145/3357384.3358028,10.1145/3397271.3401309,Su2020VL-BERT:}. In SimpleRE, we adopt BERT to model the interrelations among all possible relations in a text sequence, through BRS.

Given a sequence $X$, which contains a set of subject entities $E_s=\{E_s^1,E_s^2,\dots,E_s^n\}$, and a set of object entities $E_o = \{E_o^1, E_o^2, \dots, E_o^n\}$, we form a BRS as input to BERT: ${BRS}=\big \langle $[CLS], $X$, [SEP], $E_s^1$, [CLS], $E_o^1$, [SEP], $\dots$, [SEP], $E_s^n$, [CLS], $E_o^n$, [SEP]$\big \rangle$. [CLS] and [SEP] are the classification and separator tokens, respectively. The [CLS] tokens at different positions in the BRS input may carry different meanings, due to the different contexts.

Multiple [CLS] tokens have been used to learn hierarchical representations of a document, where one [CLS] is put in front of a sentence~\cite{liu2019fine,chen-etal-2019-group}. In BRS, multiple [CLS] tokens are for capturing different relations between entity pairs and their interrelations, because these multiple [CLS] tokens are in the same input sequence. 

%===========================
\subsection{Relation Refinement Gate}
%===========================
In BRS, representation of the first [CLS] token (denoted by $h_0$) encodes the semantic information of entire sequence. Representations of the subsequent [CLS] tokens capture the relations between each pair of entities. We denote the $i$-th relation representation as $h_{ri}$. To predict the relation type of $r_i$, in Relation Refinement Gate, we concatenate semantic representations of $h_0$ and $h_{ri}$ as $c_i=[h_0;h_{ri}]$. We then use Shallow-Deep Networks~\cite{kaya2019shallow} to compute a confidence score:
\begin{equation}
    s_c=\max\Big(\text{Sigmoid}\big(f(c_i)\big)\Big)
\end{equation}
Here $f$ denotes a single layer feed-forward neural network (FFN). If $s_c$ is larger than a predefined threshold $\tau$, $c_i$ is used to predict the target relation between $E_s^i$ and $E_o^i$, by a classifier.\footnote{We use a linear layer as a classifier for Dialogue RE and a linear layer with softmax for sentence-level RE.} Otherwise, we refine $h_0$ to be more relation-specific to $h_{ri}$, since $h_0$ is weakly related to the target relation~\cite{xue2020gdpnet}. To this end, we define a refinement mechanism to extract task-specific semantic information by updating $h_0$ for the prediction of $r_i$:
\begin{equation}\label{eqn:hp0}
    h'_{0}=\text{ReLU}\big(g(h_{ri})\big)+h_0
\end{equation}
Here $g$ is a single layer FFN and $h'_{0}$ denotes the updated semantic representation. Then $h'_{0}$ is used to predict the relation or updated further, depending on the recomputed $s_c$. To avoid possible endless refinement, we set an upper bound $B$ to limit the maximum number of iterations for refining $h'_{0}$. $B=3$ in our experiments.

\begin{table}[t]
\centering
\caption{Comparison with baselines on DialogRE. Results are 5-run averaged $F1$ with standard deviation ($\delta$).}
\begin{tabular}{l|c}
\toprule
Model & $F1\pm\delta$   \\
\midrule
CNN~\cite{yu-etal-2020-dialogue}          & 48.0$\pm$1.5   \\
LSTM~\cite{yu-etal-2020-dialogue}        & 47.4$\pm$0.6    \\
BiLSTM~\cite{yu-etal-2020-dialogue}     & 48.6$\pm$1.0    \\
\midrule
AGGCN~\cite{guo-etal-2019-attention}          & 46.2   \\
LSR~\cite{nan2020reasoning}        & 44.4    \\
DHGAT~\cite{chen2020dialogue}     & 56.1    \\
\midrule
BERT~\cite{devlin-etal-2019-bert}    & 58.5$\pm$2.0    \\
BERTs~\cite{yu-etal-2020-dialogue}    & 61.2$\pm$0.9   \\
GDPNet~\cite{xue2020gdpnet}           & 64.9$\pm$1.1    \\
SimpleRE (Ours)    & \textbf{66.3}$\pm$0.7   \\
\bottomrule
\end{tabular}

\label{tbl-dialog_eval}
\end{table}

\begin{table}
\centering
\small
\caption{Comparison with baselines on new versions of DialogRE. Results are 5-run averaged $F1$ with standard deviation.}
\begin{tabular}{l|c|c}
\toprule
Model  & English V2 ($F1\pm\delta$)  & Chinese ($F1\pm\delta$) \\ \midrule
BERT~\cite{devlin-etal-2019-bert}     & 60.6$\pm$0.5 & 61.6$\pm$0.4 \\
BERTs~\cite{yu-etal-2020-dialogue}     & 61.8$\pm$0.6 & 63.8$\pm$0.6 \\
GDPNet~\cite{xue2020gdpnet}            & 64.3$\pm$1.1  & 62.2$\pm$0.9    \\
SimpleRE (Ours)     & \textbf{66.7}$\pm$0.7   &\textbf{65.2}$\pm$1.1   \\
\bottomrule
\end{tabular}
\label{tbl-dialog_eval_2}
\end{table}

\begin{table}[t]
\centering
\small
\caption{Performance with different threshold $\tau$ values.}
\begin{tabular}{l|c c c c c c c}
\toprule
$\tau$                 &0.3 &0.4 &0.5 &0.6 &0.7 &0.8 &0.9             \\ 
\midrule
$F1$                & 67.9 &68.3 &68.0 &\textbf{69.1} &68.6 &68.1 &68.4  \\
\specialrule{.1em}{.05em}{.05em}
\end{tabular}
\label{tbl-threshold}
\end{table}

%===========================
\section{Experiments}
%===========================
We conduct experiments on Dialogue RE and sentence-level RE tasks to evaluate SimpleRE against baseline models.

%===========================
\subsection{Dataset}
%===========================
DialogRE is the first human-annotated Dialogue RE dataset~\cite{yu-etal-2020-dialogue} originated from the transcripts of American comedy ``Friends''. It contains $1,788$ dialogues and $36$ predefined relation types (see example in Table~\ref{tbl-dialog-example}). Recently, \cite{yu-etal-2020-dialogue} released a modified English version and a Chinese version of DialogRE. We evaluate SimpleRE on all three versions. 

TACRED is a widely-used sentence-level RE dataset. It contains more than $106K$ sentences drawn from the yearly TACKBP4 challenge, and has $42$ different relations (including a special ``no relation'' type). We evaluate SimpleRE on both TACRED and TACRED-Revisit (TACREV) datasets; TACREV is a modified version of TACRED. 

%===========================
\subsection{Experimental Settings}
%===========================
We compare SimpleRE with two recent BERT-based methods, BERTs~\cite{yu-etal-2020-dialogue} and GDPNet~\cite{xue2020gdpnet}. We also include popular baselines  AGGCN~\cite{guo-etal-2019-attention}, LSR~\cite{nan2020reasoning}, and DHGAT~\cite{chen2020dialogue} in our experiments. For a fair comparison with BERTs and GDPNet, we utilize the same hyperparameter settings, except for batch size. Specifically, we set batch size to $6$ rather than $24$ for SimpleRE because it predicts multiple relations (\ie all relations annotated in one dialogue) per forward process. To set threshold $\tau$, we conduct a preliminary study on the development set with different $\tau$ values. Reported in Table~\ref{tbl-threshold}, SimpleRE achieves best performance when $\tau\approx 0.6$. Thus, we set $\tau=0.6$ throughout the experiments, unless specified otherwise. We set the maximal refinement iterations $B$ as $3$ for Dialogue RE. 

\begin{table}[t]
\centering
\small
\caption{Average training time (in minutes)  per epoch on Dialogue RE}
\begin{tabular}{l|c}
\toprule
Model                 & Average Time (mins)             \\ 
\midrule
BERT \cite{devlin-etal-2019-bert}                & 4.7  \\
BERTs \cite{yu-etal-2020-dialogue}     & 4.7     \\
GDPNet \cite{xue2020gdpnet}      & 12.6     \\
SimpleRE (Ours)               & 0.9      \\
\bottomrule
\end{tabular}

\label{tbl-dialogre_time}
\end{table}

\begin{table}[t]
\centering
\caption{Ablation study of SimpleRE on DialogRE.}
\begin{tabular}{p{1.5in}|c c}
\toprule
Model                 & $F1\pm \sigma$              \\ 
\midrule
SimpleRE                & \textbf{66.3}$\pm$0.7  \\
SimpleRE w/o BRS      & 60.4$\pm$0.9     \\
SimpleRE w/ BRS-v2      & 62.8$\pm$1.1     \\
SimpleRE w/ BRS-v3      & 63.5$\pm$0.8     \\
SimpleRE w/o RRG              & 65.5$\pm$0.7      \\
\specialrule{.1em}{.05em}{.05em}
\end{tabular}
\label{tbl-dialogre_a_study}
\end{table}

%===========================
\subsection{Results on DialogRE}
%===========================

%===========================
\subsubsection{Performance by $F1$.} 
%===========================
Table~\ref{tbl-dialog_eval} summarizes the results on DialogRE. Observe that BERT-based models significantly outperform non-BERT models. Among the three BERT-based models, SimpleRE surpasses GDPNet and BERTs by $1.4\%$ and $5.1\%$ respectively, on DialogRE, by $F1$ measure. 

Note that, the two modified DialogRE datasets, English V2 and Chinese version, are released recently. Since most existing models have not reported their performance, we obtain results by running author released codes of existing BERT-based models. Table~\ref{tbl-dialog_eval_2} shows that SimpleRE achieves better performance than baselines on both versions of DialogRE.

%===========================
\subsubsection{Efficiency by training time.} 
%===========================
The average training time per epoch is reported in Table~\ref{tbl-dialogre_time}, after training for $20$ epochs. SimpleRE is about $5\times$ faster than baselines, despite its smaller batch size. Note a dialogue in Dialogue RE may contain multiple relations. Existing models only infer one relation per forward process. On the contrary, SimpleRE predicts multiple relations per forward process. Its simple structure leads to better efficiency than baselines, \eg GDPNet with SoftDTW~\cite{cuturi2017soft}.

%===========================
\subsubsection{Ablation study.} 
%===========================
We conduct ablation studies on DialogRE, for the effectiveness of components in SimpleRE: BERT Relation Token Sequence (BRS) and Relation Refinement Gate (RRG). To evaluate their impacts, we remove BRS and RRG from our model separately. To remove BRS, we change the input format to predict one relation each time with a modified input format: $\langle$[CLS], $X$, [SEP], $E_s$, [CLS], $E_o$, [SEP]$\rangle$. To remove RRG, all relations are predicted based on the corresponding token representations $[h_0;h_{ri}]$, without updating $h_0$. Besides, we also design two alternative BRS, \ie BRS-v2 and BRS-v3, for comparison. For BRS-v2, we exchange the [CLS] tokens between entity pairs with [SEP] tokens before subject entities. Similarly, we exchange the [CLS] tokens with [SEP] tokens after object entities in BRS-v3.

Reported in Table~\ref{tbl-dialogre_a_study}, the results show that removing BRS leads to large performance degradation, indicating interrelations among relations have a significant impact on RE performance. Meanwhile, RRG module also contributes to the performance gains. Another interesting finding is that BRS-v2 and -v3 cannot model the relations well although they both use multiple [CLS] tokens in the input sequence. The $F1$ score decreases from $66.3$ to $62.8$ and $63.5$, respectively. This result further shows that [CLS] token in BRS is sensitive to its position in the sequence.

\begin{table}[t]
\centering
\small
\setlength{\tabcolsep}{4.0 pt}
\caption{$F1$ of all models on TACRED and TACRED-Revisit (TARREV), the sentence-level RE datasets.}
\begin{tabular}{l|l r}
\toprule
Model &  TACRED & TACREV \\
\midrule
LSTM~\cite{zhang-etal-2017-position}     & 62.7 & 70.6 \\
PA-LSTM~\cite{zhang-etal-2017-position}  & 65.1 & 74.3 \\
C-AGGCN~\cite{guo-etal-2019-attention}   & 68.2 & 75.5 \\
LST-AGCN~\cite{sun2020relation}          & 68.8 & -    \\ 
\midrule
SpanBERT~\cite{joshi2020spanbert}        & 70.8 & 78.0 \\
GDPNet~\cite{xue2020gdpnet}              & 70.5 & 80.2 \\
SimpleRE (Ours)                          & \textbf{71.7}  & \textbf{80.7} \\
\midrule
KnowBERT~\cite{Peters2019KnowledgeEC}    & 71.5 & 79.3  \\
\bottomrule
\end{tabular}
\label{tbl-tacred_eval}
\end{table}

\begin{table}[t]
\small
\centering
\caption{Ablation study of SimpleRE on TACRED. For the model without $2^{\text{nd}}$ [CLS] token, we use [SEP] token. Instead of using [$h_0:h_r$], we have evaluated SimpleRE with either $h_0$ or $h_r$ alone for relation prediction.}
\begin{tabular}{l|c c}
\toprule
Model                 & $F1$              \\ 
\midrule
SimpleRE                & \textbf{71.7}  \\
SimpleRE w/o $2^{\text{nd}}$ [CLS] token     & 70.6     \\
SimpleRE w/o relation representation $h_{r}$              & 70.0      \\
SimpleRE w/o semantic representation $h_{0}$              & 70.6      \\
\bottomrule
\end{tabular}
\label{tbl-tacred_a_study}
\end{table}

%===========================
\subsection{Results on TACRED}
%===========================
We now adapt SimpleRE to sentence-level RE. Note that SimpleRE was not evaluated on document-level RE due to the difference in problem settings. We leave the adaptation of SimpleRE to document-level RE as our future work. 

Because each sentence only contains a single relation in  sentence-level RE dataset, BRS becomes $\langle$[CLS], $X$, [SEP], $E_s$, [CLS], $E_o$, [SEP] $\rangle$. The representations of the two [CLS] tokens are concatenated for relation prediction. Compared to typical RE input sequence $\langle$[CLS], $X$, [SEP], $E_s$, [SEP], $E_o$, [SEP] $\rangle$, SimpleRE replaces the [SEP] token between two entities with a [CLS] token. RRG is not applicable here because there is only one relation in each sentence. Hence, it is unnecessary to refine $h_0$ to be target relation specific. 

For fair comparison, we refer~\cite{xue2020gdpnet} to use SpanBERT as the backbone (\ie BERT in Figure~\ref{fig:overview}). Results on TACRED and TACREV are summarized in Table~\ref{tbl-tacred_eval}. SimpleRE outperforms all baselines including KnowBERT~\cite{Peters2019KnowledgeEC} on both datasets. Note that KnowBERT incorporates external knowledge base during training. 

Table~\ref{tbl-tacred_a_study} summarizes the results of ablation studies on TACRED. We first replace the second [CLS] token with a [SEP] token, which leads to $1.1\%$ performance degradation. This result suggests that [CLS] is necessary to capture the relation between entities near it. Moreover, the performance of our model further drops without relation representation $h_r$, \ie predicting relation purely based on $h_0$ instead of $[h_0:h_r]$. Poorer performance is also observed when only $h_r$ is used for prediction. Hence, both $h_0$ and $h_r$ contribute to the correct prediction of relations. 

In short, through experiments on both dialogue and sentence-level RE tasks, we show SimpleRE is a strong competitor. Although simple, both components, BRS and RRG, are essential in SimpleRE's model design. 

%===========================
\section{Conclusion}
%===========================
In this paper, we propose a simple yet effective model for dialogue relation extraction. Building on top of the powerful modeling capability of BERT, SimpleRE is designed to learn and reason the interrelations among multiple relations in a dialogue. The most important component in SimpleRE is the BERT Relation Token Sequence, where multiple [CLS] tokens are used to capture relations between entity pairs. The Relation Refinement Gate is designed to further improve the semantic representation in an adaptive manner. Through experiments and ablation studies, we show that both components contribute to the success of SimpleRE. Due to its simple structure and fast training speed, we believe SimpleRE serves a good baseline in Dialogue RE task. The SimpleRE can also be easily adapted to sentence-level relation extraction. On datasets for both tasks, DialogRE and TACRED, we show that our simple model is a strong competitor for relation extraction tasks. 

\vfill\pagebreak

\bibliographystyle{IEEEbib}
\bibliography{main}

\end{document}